\documentclass[conference]{IEEEtran}
\IEEEoverridecommandlockouts
\usepackage{cite}
\usepackage{amsmath,amssymb,amsfonts}
\usepackage{algorithmic}
\usepackage{graphicx}
\usepackage{textcomp}
\usepackage{comment}
\usepackage{xcolor}
\usepackage{multirow}
\usepackage{url}
\usepackage{subcaption}
\usepackage{threeparttable}
\usepackage{tabularx,array,booktabs}
\def\BibTeX{{\rm B\kern-.05em{\sc i\kern-.025em b}\kern-.08em
    T\kern-.1667em\lower.7ex\hbox{E}\kern-.125emX}}
\begin{document}

\title{Towards reconstructing experimental sparse-view X-ray CT data with diffusion models\\

\thanks{This work was supported by the Dutch Research Council (NWO, file number VI.Vidi.223.059)}
}

\author{
\IEEEauthorblockN{Nelas J. Thomsen\IEEEauthorrefmark{1}\IEEEauthorrefmark{2}, 
Xinyuan Wang\IEEEauthorrefmark{2}, 
Felix Lucka\IEEEauthorrefmark{2}, 
Ezgi Demircan-Tureyen\IEEEauthorrefmark{2}}
\IEEEauthorblockA{\IEEEauthorrefmark{1}Martin-Luther-University Halle-Wittenberg, Institute of Physics, Halle, Germany}
\IEEEauthorblockA{\IEEEauthorrefmark{2}Centrum Wiskunde \& Informatica, Computational Imaging Group, Amsterdam, The Netherlands}
}

\maketitle

\begin{abstract}
    Diffusion-based image generators are promising priors for ill-posed inverse problems like sparse-view X-ray Computed Tomography (CT). As most studies consider synthetic data, it is not clear whether training data mismatch (``domain shift'') or forward model mismatch complicate their successful application to experimental data. We measured CT data from a physical phantom resembling the synthetic Shepp-Logan phantom and trained diffusion priors on synthetic image data sets with different degrees of domain shift towards it. Then, we employed the priors in a Decomposed Diffusion Sampling scheme on sparse-view CT data sets with increasing difficulty leading to the experimental data. Our results reveal that domain shift plays a nuanced role: while severe mismatch causes model collapse and hallucinations, diverse priors match or exceed well-matched but narrow priors. Forward model mismatch pulls the image samples away from the prior manifold, which causes artifacts but can be mitigated with annealed likelihood weight schedules that also increase computational efficiency. Overall, we demonstrate that performance gains do not immediately translate from synthetic to experimental data, and future development must validate against real-world benchmarks.
    
\end{abstract}


\section{Introduction}
Motivated by the need to reduce radiation and scan times, sparse-view X-ray CT presents an ill-posed inverse problem historically addressed by model-based \cite{sidky2008image, chen2008prior}, and learned \cite{jin2017deep, chen2018learn, adler2018learned} methods. More recently, diffusion-based image generators are increasingly used as plug-and-play solvers \cite{songsolving, chung2022improving, chung2023diffusion, zhang2025improving, chung2024decomposed}. These methods hinge on balancing two competing forces: the prior (what the diffusion model says a typical image should look like) and the likelihood (what the CT measurement data requires the image to be). The interplay between these two components is critical, as excessive reliance on the prior can lead to loss of detail or hallucinations, while overemphasizing data fidelity may amplify artifacts or fail to resolve ambiguities arising from the ill-posedness of the inverse problem.  

The balance becomes critical for experimental data, where the gap between model and reality widens. The likelihood is challenged by the mismatch between the simplified forward model and the physics of real-world acquisition. While simulated measurements follow exact mathematical projectors, experimental scans include geometric inaccuracies, beam hardening, scatter, and complex noise textures that the standard likelihood term often cannot capture. As a result, the solver may either amplify artifacts by fitting the wrong data model or rely too heavily on the prior to mask the mismatch.

The challenge in the prior, on the other hand, stems from the distribution shift between the training data and the experimental target. Standard diffusion models are often trained on idealized synthetic datasets that lack the geometric and textural variation of real objects. Priors trained on such narrow distributions can become brittle on experimental data, failing to reconstruct features that fall outside the training set. Conversely, priors specialized to experimental geometries may generalize poorly to unseen variations. Robust experimental reconstruction therefore requires not only handling forward model mismatch but also designing priors that bridge the gap between clean synthetic and imperfect experimental domains.

In this work, we examine how the performance of diffusion-based solvers for sparse-view CT translates to real experimental settings. To enable a controlled comparison, we designed a physical phantom that resembles the standard Shepp-Logan phantom, isolating specific sources of error. Our contributions are threefold: First, we analyze the impact of training distribution and demonstrate that while rigid geometric mismatch leads to drastic failure, a diverse prior may offer superior robustness against geometric challenges inherent to real-world objects. Second, we investigate forward model mismatch, showing that it manifests as not only severe streak artifacts, but also as structural misalignments, a specific form of hallucination where the solver distorts features to fit physical inconsistencies. Finally, we address practical inference strategies, demonstrating that an annealed likelihood weight schedule can reconciles the diffusion prior with imperfect experimental measurements even with significantly fewer sampling steps. 

\section{Related Work}
\subsection{Preliminaries on Diffusion Models} 
Diffusion models \cite{ho2020denoising} generate images by learning to invert a predefined noising process. In denoising diffusion probabilistic models (DDPMs) \cite{ho2020denoising}, this inversion is formulated as a Markovian reverse-time process, where each denoising step is modeled as a Gaussian distribution parameterized by a neural network $\epsilon_\theta$. The network is trained to predict the noise realization at a given state, which effectively estimates the mean of the reverse conditional distribution. Deterministic sampling schemes like DDIM \cite{song2020denoising} leverage this to define a non-Markovian trajectory, allowing for the direct estimation of the clean image $\hat{\mathbf{x}}_t$ from a noisy observation $\mathbf{x}_t$ at any timestep. This estimate, often referred to as the Tweedie denoised estimate \cite{efron2011tweedie}, is given by $\hat{\mathbf{x}}_t := \frac{1}{\sqrt{\bar{\alpha}_t}}
\left(\mathbf{x}_t - \sqrt{1-\bar{\alpha}_t}\,\epsilon^{(t)}_{\theta^*}(\mathbf{x}_t)\right)$, where $\bar{\alpha}_t$ is the noise schedule parameter and $\epsilon_{\theta}$ is the predicted noise.
\subsection{Diffusion-based Priors for Tomography} 
Assuming the tomographic measurement model $\mathbf{y} = \mathbf{A}\mathbf{x}_0 + \mathbf{n}$, the goal is to reconstruct the unknown image $\mathbf{x}_0 \in \mathbb{R}^{n}$ from projection data $\mathbf{y} \in \mathbb{R}^{m}$, where $\mathbf{n} \in \mathbb{R}^{m}$ models measurement noise. Here $\mathbf{A}:\mathbb{R}^{n}\to\mathbb{R}^{m}$ denotes the known tomography operator such as a discretized Radon transform mapping an image to line-integral measurements. By Bayes' Rule, the posterior score (i.e., the gradient of the log-posterior) decomposes as $\nabla_{\mathbf{x}_t}\log p(\mathbf{x}\mid \mathbf{y})
=
\nabla_{\mathbf{x}_t}\log p(\mathbf{x})
+
\nabla_{\mathbf{x}_t}\log p(\mathbf{y}\mid \mathbf{x})$
with the score $\nabla_{\mathbf{x}_t}\log p(\mathbf{x})$ being approximated by a pretrained model $\epsilon_\theta(\mathbf{x}_t,t)$. Thus, posterior sampling can be seen as combining a learned “prior direction” 
$\epsilon_\theta(\mathbf{x}_t,t)$ with a physics-based “data direction” 
$\mathbf{A}^\top(\mathbf{y}-\mathbf{A}\mathbf{x})$, assuming $\mathbf{n} \sim \mathcal{N}(\mathbf{0}, \sigma_y^2 \mathbf{I})$. Sampling approaches vary in the way they combine these two directions. Plug-and-play (PnP) approaches operate by injecting a likelihood-guidance term into each reverse diffusion step \cite{songsolving, chung2022improving, chung2023diffusion, chung2024decomposed, zhang2025improving}. Given the data consistency loss $\ell(\mathbf{x}) = \frac{1}{2}\,\lVert \mathbf{y} - \mathbf{A}\mathbf{x} \rVert^2$, the widely used diffusion posterior sampler (DPS) \cite{chung2023diffusion} updates the iterate at each reverse diffusion step according to:
\begin{equation}
\label{eq:reverse_update}
\mathbf{x}_{t-1}
= \sqrt{\bar{\alpha}_{t-1}}
\left(\hat{\mathbf{x}}_t - \gamma_t \nabla_{\mathbf{x}_t}\,\ell\!\left(\hat{\mathbf{x}}_t\right)\right)
+ \tilde{\mathbf{w}}_t(\eta)
\end{equation}
where \(\gamma_t\) denotes the data-consistency step size and \(\tilde{\mathbf{w}}_t(\eta)\) is the overall noise term controlled by \(\eta \in [0,1]\), interpolating between fully deterministic and fully stochastic ancestral sampling. Decomposed Diffusion Sampler (DDS)~\cite{chung2024decomposed} replaces the computationally expensive Jacobian term in Eq.~\eqref{eq:reverse_update} with a gradient-based surrogate, converting a step on the noisy generative manifold into a projected step onto the clean image manifold \(\mathcal{M}\), i.e., $\hat{\mathbf{x}}_t - \gamma_t \nabla_{\mathbf{x}_t}\,\ell\!\left(\hat{\mathbf{x}}_t\right)
= \mathcal{P}_{\mathcal{M}}\!\left(\hat{\mathbf{x}}_t - \zeta_t \nabla_{\hat{\mathbf{x}}_t}\,\ell\!\left(\hat{\mathbf{x}}_t\right)\right)$. This projection is then implemented via an $M$-step conjugate gradient (CG), i.e., $\mathrm{CG}\!\left(\mathbf{A}^\top \mathbf{A},\; \mathbf{A}^\top \mathbf{y},\; \hat{\mathbf{x}}_t,\; M\right)$.
 
\section{Methodology}

To realize an experimental scan of an object resembling the canonical Shepp-Logan (SL) phantom, we designed a physical phantom by laser-cutting ellipses into a 6mm-thick transparent polymethyl methacrylate (PMMA) plate. Two ellipses were left air-filled, while the others were filled with deep-pour transparent epoxy resin (see supplementary Fig.~S1 for a photograph). Scans were acquired on a custom-built laboratory X-ray CT scanner in the FleX-ray Lab at CWI in Amsterdam.We collected 901 equally spaced projections over a full rotation in cone-beam geometry, with a source-to-object distance of 234.92 mm and a source-to-detector distance of 400 mm. The X-ray tube operated at 70 kVp and 600 $\mu A$ with a 30 ms exposure. Projections were recorded with an effective detector pixel size of 0.2992 mm and 478 detector channels per view. Only the central detector row was used, yielding a fan-beam-equivalent sinogram of size 901×478 (views $\times$ channels). Dark-field and flat-field references were acquired for correction.
   
\subsection{Training Data}
Let a phantom be defined as a set of ellipses \(E=\{e_i\}_{i=1}^{m}\), where each ellipse
\(e_i := (A_i,a_i,b_i,x_{0,i},y_{0,i},\phi_i)\) encodes intensity, semi-axes, center coordinates, and orientation.
For each generated phantom, we construct a sampled set \(E\) by perturbing a mean set \(E_\mu=\{e_{i,\mu}\}_{i=1}^{m}\).
Specifically, for each ellipse \(i\) and each parameter \(p\in\{A,a,b,x_0,y_0,\phi\}\), we sample $
p_i = p_{i,\mu} + s_p\,\varepsilon$, with $\varepsilon\sim\mathcal{N}(0,\sigma_p^2)$
where \(s_p\) is the corresponding scale, i.e.,
\[
s_p=
\begin{cases}
|p_{i,\mu}|, & p\in\{A,a,b\},\\
\text{image\_width}, \text{image\_height}, & p=x_0, p=y_0,\\
1, & p=\phi.
\end{cases}
\]
and \(\sigma_p\) is listed in Table~\ref{tab:sigma_params}. Additionally, each ellipse is randomly dropped, or a new random ellipse is added, with probability $\boldsymbol{\textit{p}_{\mathrm{add,drop}}}$, also listed in the table. This way, we have created three training sets, each of which has 10K images:

\begin{itemize}
    \item \textit{Standard Shepp-Logan dataset} ($\mathcal{X}_\text{std}$): For each phantom, we set \(E_\mu := E_{\mathrm{std}}\), where $E_{\mathrm{std}}$ is the set of ellipses from the standard Shepp-Logan test image, and sample \(E\) by perturbing all ellipses as above.
    \item \textit{Experimental Shepp-Logan dataset} 
    ($\mathcal{X}_\text{exp}$):
    For each phantom, we set \(E_\mu := E_{\mathrm{exp}}\), where $E_{\mathrm{exp}}$ is the set of ellipses from our laser-cut experimental SL, and sample \(E\) by perturbing all ellipses as above.
    \item \textit{Mixed Shepp-Logan dataset} ($\mathcal{X}_\text{mix}$): For each phantom, we first sample \(Z \sim \mathrm{Bernoulli}(\pi)\). If \(Z=1\), we set \(E_\mu := E_{\mathrm{exp}}\); otherwise \(E_\mu := E_{\mathrm{std}}\). We then sample \(E\) by perturbing all ellipses as above. Here \(\pi\in[0,1]\) controls the fraction of experimental- versus SL-centered samples in the mixed dataset.
\end{itemize}

We refer the reader to supplementary Fig. S2 for visual examples of the generated phantoms from these datasets.

\begin{table}[t]
\centering
\begin{threeparttable}
\caption{Parameters Used for Tailoring Training Sets}
\label{tab:sigma_params}
\footnotesize
\setlength{\tabcolsep}{5pt}
\renewcommand{\arraystretch}{1.0}
\begin{tabular}{l c c c c c c}
\toprule
\textbf{Dataset} & \textbf{Center} & $\boldsymbol{\sigma_A}$ & $\boldsymbol{\sigma_{a,b}}$ & $\boldsymbol{\sigma_{x_0,y_0}}$ & $\boldsymbol{\sigma_\phi}$ & $\boldsymbol{\textit{p}_{\mathrm{add,drop}}}$ \\
\midrule
$\mathcal{X}_\text{exp}$ & $E_{\mathrm{exp}}$ & 0.03 & 0.02 & 0.01 & $10^\circ$ & 0.01 \\
$\mathcal{X}_\text{std}$ & $E_{\mathrm{std}}$ & 0.03 & 0.02 & 0.01 & $10^\circ$ & 0.01 \\
$\mathcal{X}_\text{mix}$ & $E_{\mathrm{mix}}^{\mathrm{a}}$ & 0.03 & 0.03 & 0.02 & $45^\circ$ & 0.03 \\
\bottomrule
\end{tabular}
\begin{tablenotes}
\scriptsize
\item[a] $E_{\mathrm{mix}} = \pi E_{\mathrm{exp}} + (1-\pi)E_{\mathrm{std}}$
\end{tablenotes}
\end{threeparttable}
\vspace{-8pt}
\end{table}

\subsection{Generative Priors}
Using these datasets, we train three corresponding priors. The \textit{Standard SL prior} ($\mathbf{f}_\text{std}$) models the narrow distribution of ideal canonical phantoms. The \textit{Experimental SL prior} ($\mathbf{f}_\text{exp}$) is explicitly tailored to the geometric design of our physical laser-cut phantom. Finally, the \textit{Mixed SL prior} ($\mathbf{f}_\text{mix}$) provides a broader distribution by encompassing both standard and experimental variations.

\subsection{Reconstruction Pipeline} 
Following Section II-B, we use DDS for reverse diffusion sampling from random noise with a pre-trained model (guided-diffusion codebase\footnote{https://github.com/openai/guided-diffusion}), guided by measurement-based CG steps. To match the real setup, the forward operator $\mathbf{A}$ models a fan-beam geometry with equally spaced sparse projections over [0°,180°]. Unless stated otherwise, we reconstruct $128 \times 128$ images and use a cosine noise schedule \cite{nichol2021improved} to preserve structure across timesteps at low resolution. In Sections IV-A and IV-B, we fix the number of model evaluations (NFE) to 1000, $M=5$, $\eta=0.85$, $\sigma_y=10^{-7}$, and $\gamma_t=1 ; \forall t$.

\subsection{Test domains}
We evaluate our models across four progressively challenging domains. The \textit{Standard simulation} ($\mathbf{y}_\text{sim(std)}$) is synthesized directly from the canonical SL image. The \textit{CAD simulation} ($\mathbf{y}_\text{sim(cad)}$) is generated from the laser-cut design parameters $E_{\mathrm{exp}}$, mimicking standard piecewise geometry but with reduced contrast. The \textit{Recon. simulation} ($\mathbf{y}_\text{sim(recon)}$) introduces realistic textures by forward-projecting from a full-view reconstruction of the physical phantom. Finally, the \textit{Experimental measurement} ($\mathbf{y}_\text{exp}$) is acquired directly from the physical phantom setup.

\section{Experimental Results}
\subsection{Impact of Training Distribution}
\begin{figure*}[t!]
 \centering
 \includegraphics[width=\linewidth]{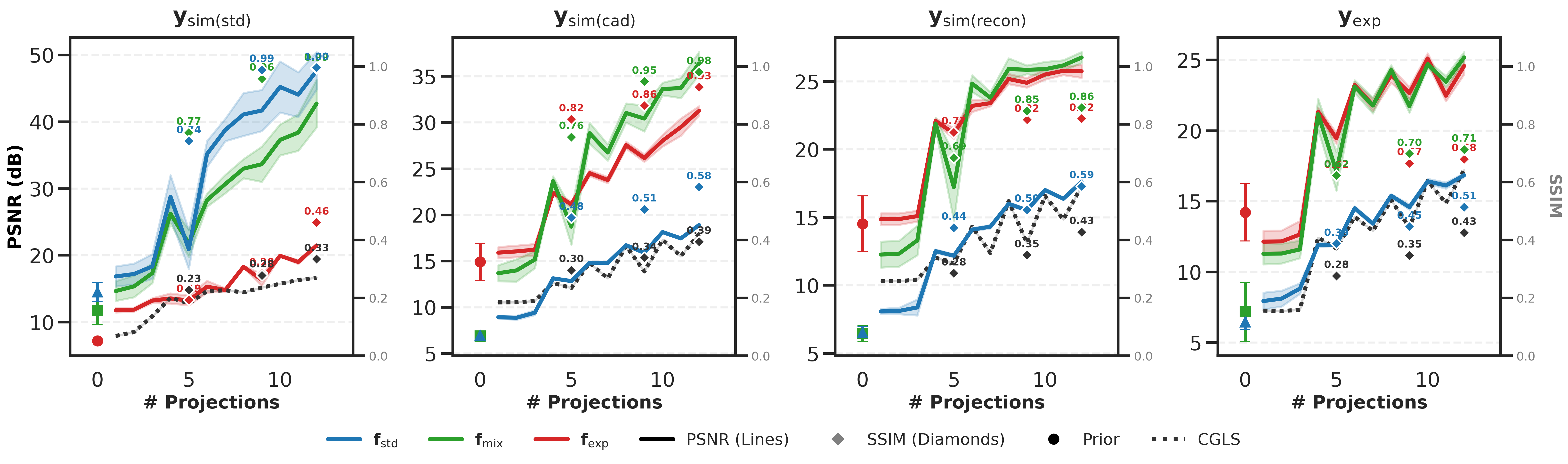}
  \caption{PSNR (dB) as a function of the number of projections across four different test domains for the reconstructions obtained using CGLS and DDS with three different diffusion priors. Solid lines show PSNR, while diamond markers indicate sampled SSIM values at selected projection counts (5, 9, 12). Shaded areas denote $\pm1$ standard deviation across 10 random seeds.}
 \label{fig:psnrplots}
\end{figure*}

Fig.~\ref{fig:psnrplots} quantifies reconstruction performance across a representative simulation-to-experimental spectrum. In the idealized $\mathbf{y}_\text{sim(std)}$ domain (left), the domain-specific $\mathbf{f}_\text{std}$ achieves the highest PSNR, while $\mathbf{f}_\text{mix}$ shows a moderate generalization penalty and $\mathbf{f}_\text{exp}$  drops to baseline due to geometric mismatch between the standard SL and experimental laser-cut phantoms. In the second panel, where the test domain ($\mathbf{y}_\text{sim(cad)}$) introduces a geometric shift toward the experimental phantom, $\mathbf{f}_\text{mix}$ unexpectedly exceeds the domain-specific $\mathbf{f}_\text{exp}$ by about 5 dB for more than five projections, possibly because its exposure to high-contrast standard phantoms provides stronger edge cues that transfer to boundary preservation despite the contrast difference. In the final two panels, the reference shifts from idealized phantoms to full-view reconstructions of experimental data, linking evaluation more directly to achievable experimental quality. These panels differ only in the forward model: $\mathbf{y}_\text{sim(recon)}$ uses simulated projections from the reconstructed volume, whereas $\mathbf{y}_\text{exp}$ uses real measurements. In both cases, the PSNR gap between $\mathbf{f}_\text{mix}$ and the domain-specific $\mathbf{f}_\text{exp}$ largely closes across projection angles. However, the fully experimental case ($\mathbf{y}_\text{exp}$) shows consistently lower absolute PSNR than ($\mathbf{y}_\text{sim(recon)}$), isolating the additional impact of non-ideal forward-model effects.

\begin{figure}[t!]
 \centering
 \includegraphics[width=0.92\linewidth]{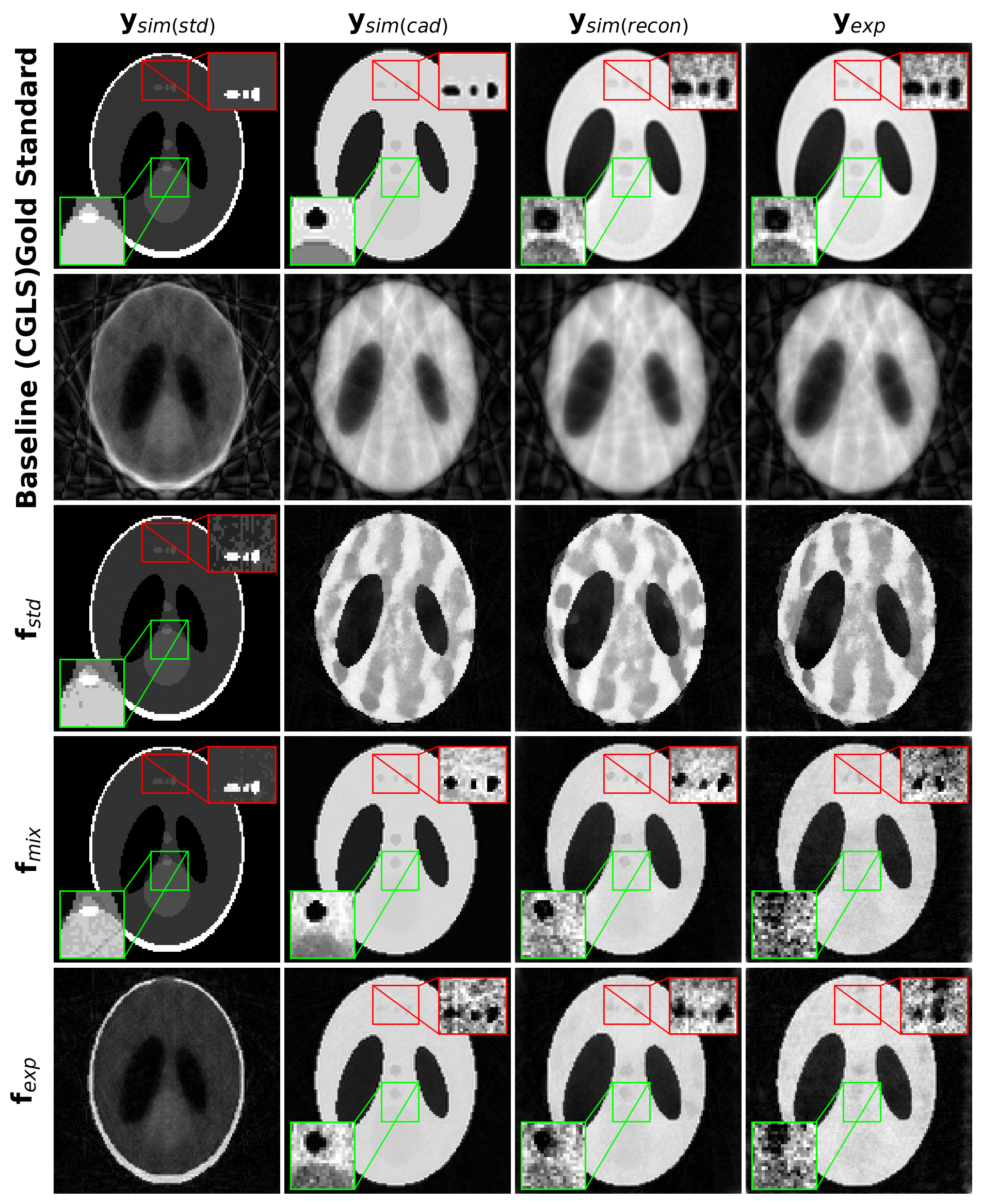}
 \caption{Example reconstructions (averaged over nine samples) from 12 sparse projections across four test domains (rows). From top to bottom: gold standard, baseline without a prior, and reconstructions using the standard SL ($\mathbf{f}_\text{std}$), mixed ($\mathbf{f}_\text{mix}$), and experimental SL ($\mathbf{f}_\text{exp}$) diffusion priors.}
 \label{fig:recons}
\end{figure}

Fig.~\ref{fig:recons} shows reconstructions from 12 sparse projections across four domains. Baseline reconstructions with Conjugate Gradient Least Squares (CGLS) confirm that the measurements alone are insufficient. While diffusion priors generally recover high-fidelity structures, large distribution shifts (such as applying \(\mathbf{f}_\text{std}\) to the laser-cut phantom or \(\mathbf{f}_\text{exp}\) to the standard SL) lead to catastrophic failure. In contrast, the mixed prior \(\mathbf{f}_\text{mix}\) generalizes robustly, reconstructing all domains with performance comparable to specialized priors on their native distributions. The narrower priors, \(\mathbf{f}_\text{std}\) and \(\mathbf{f}_\text{exp}\), can still leave residual streaks even on in-distribution samples, suggesting that the solver struggles to balance prior adherence with data consistency. The broader support of \(\mathbf{f}_\text{mix}\) appears to suppress these streaks, indicating that a looser prior may better accommodate the likelihood term and help the diffusion process find natural solutions without overfitting high-frequency inconsistencies.

\subsection{Impact of Forward Model Mismatch }
\begin{figure}[t!]
 \centering
 \includegraphics[width=\linewidth]{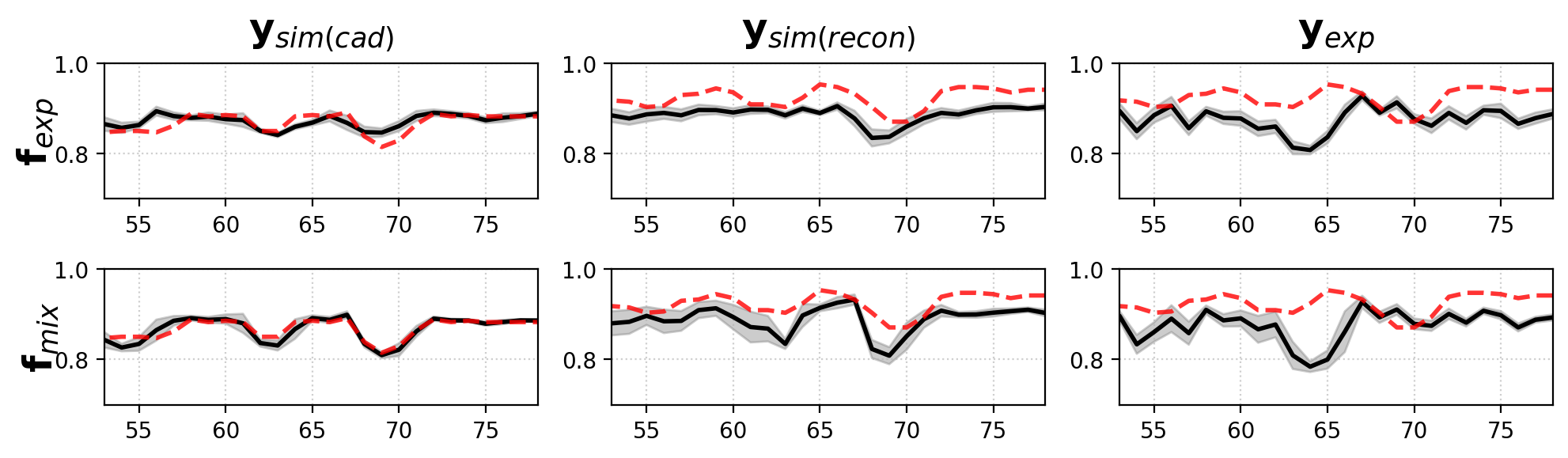}
  \caption{Line profiles from Fig.~\ref{fig:recons} through the three small holes at row 25 of the laser-cut phantom. Results involving $\mathbf{f}_\text{std}$ and/or $\mathbf{y}_\text{sim(cad)}$ are omitted. The gold standard is shown as a red dashed line, and shaded regions indicate $\pm1$ standard deviation over 10 random seeds.}
 \label{fig:lineprofiles}
\end{figure}

As noted above, the performance gap between \(\mathbf{y}_\text{sim(recon)}\) and \(\mathbf{y}_\text{exp}\) is attributable to forward model mismatch. Fig.~\ref{fig:lineprofiles} illustrates this through line profiles from row 25 of the laser-cut phantom, which contains three fine holes representing its highest-frequency details. In the \(\mathbf{y}_\text{sim(cad)}\) domain, especially with \(\mathbf{f}_\text{mix}\), the profiles closely follow the ground truth. In contrast, experimental data introduce a systematic offset. In the fully experimental domain, \(\mathbf{y}_\text{exp}\), the mismatch causes the likelihood term to dominate the sampling process and pull the trajectory away from the clean prior manifold. This increases the risk of hallucinations, appearing here as a spatial shift of the reconstructed profile dips relative to the ground truth. The misalignment suggests that, to satisfy data consistency under a mismatched physics model, the solver not only propagates artifacts but also distorts features.

\begin{figure}[t!]
 \centering
 \includegraphics[width=0.9\linewidth]{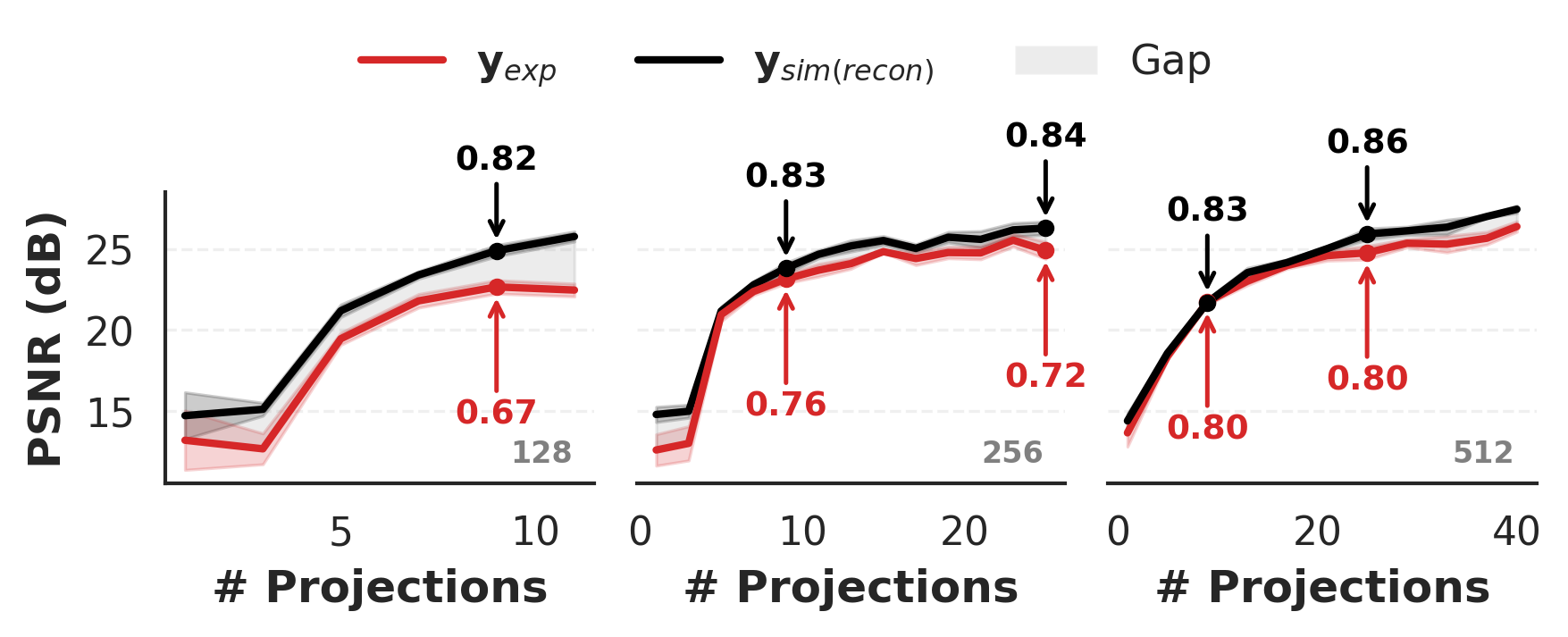}
  \caption{PSNR (dB) vs number of projections for three resolutions (128, 256, 512) using $\mathbf{f}_\text{exp}$ on experimental data $\mathbf{y}_\text{exp}$ and simulated data $\mathbf{y}_\text{sim(recon)}$, revealing the performance gap caused by forward model mismatch. Arrows indicate SSIM values at selected projection counts (9, 25). Shaded bands denote $\pm 1$ standard deviation.}
 \label{fig:higherresolutions}
\end{figure}

Since discretization errors typically decrease at finer grids, we tested whether higher reconstruction resolution could reduce the performance gap. Fig.~\ref{fig:higherresolutions} shows results at \(128\), \(256\), and \(512\) resolution. Although the gap between experimental (\(\mathbf{y}_\text{exp}\)) and simulated (\(\mathbf{y}_\text{sim(recon)}\)) reconstructions narrows as resolution increases, a systematic deficit remains even at \(512 \times 512\) (See supplementary Fig.~S3 for reconstruction examples and corresponding error maps.) This persistent gap confirms that physical model mismatch, such as spectral inconsistency and geometric calibration error, remains a major source of experimental CT error.

\subsection{Impact of Inference Design}
In DDS, the trade-off between the generative prior and measured data is controlled by the number of CG iterations and the likelihood weight \(\gamma\), introduced in \cite{chung2024decomposed} to modulate the measurement constraint. This balance is enforced in the proximal update by minimizing
\begin{equation}
\label{eq:reg_loss}
L(x) = \frac{\gamma}{2}\ell(\mathbf{x}) + \frac{1}{2}\|\mathbf{x} - \hat{\mathbf{x}}_t\|_2^2,
\end{equation}
which yields
\[
\mathrm{CG}\!\left(\gamma\,\mathbf{A}^\top \mathbf{A} + \mathbf{I},\; \hat{\mathbf{x}}_t + \gamma\,\mathbf{A}^\top \mathbf{y},\; \hat{\mathbf{x}}_t,\; M\right).
\]
Here, \(\gamma\) scales the likelihood relative to the proximity term imposed by the diffusion prior.

\begin{table*}[t!]
    \centering
    \newcommand{\bms}[2]{$\mathbf{#1}_{\scriptscriptstyle #2}$}
    \newcommand{\ms}[2]{$#1_{\scriptscriptstyle #2}$}
    \caption{Quantitative comparison (PSNR / SSIM) of constant versus linear decay likelihood weight scheduling across different NFEs and view counts at a $256 \times 256$ resolution, for both $\mathbf{y}_\text{sim(recon)}$ and $\mathbf{y}_\text{exp}$. Subscripts indicate standard deviations across nine runs. Bold values indicate the best mean results within each column for a given NFE group. Const: Constant schedule; Linear: Linear Decay schedule ($\gamma_m$ denotes $\gamma_\text{max}$). See Supplementary Figs.~S4 and S5 for visual samples from the diffusion trajectory, and Supplementary Table~T1 for an additional row with further reduced sampling steps (NFE=10).}
    \label{tab:scheduling_results}
    \setlength{\tabcolsep}{3pt} 
    \begin{tabular}{c l c c c c}
        \toprule
        & & \multicolumn{2}{c}{\textbf{12 views}} & \multicolumn{2}{c}{\textbf{24 views}} \\
        \cmidrule(lr){3-4} \cmidrule(lr){5-6}
        \textbf{NFE} & \textbf{Schedule} & $\mathbf{y}_\text{sim(recon)}$ & $\mathbf{y}_\text{exp}$ & $\mathbf{y}_\text{sim(recon)}$ & $\mathbf{y}_\text{exp}$ \\
        \midrule
        \multirow{5}{*}{1000} 
          & Const ($\gamma=0.5$)    &  \ms{24.35}{0.29} / \ms{0.831}{0.001} & \ms{24.07}{0.21} / \ms{0.774}{0.002} &  \ms{25.94}{0.30} / \ms{0.851}{0.001} & \ms{24.91}{0.45} / \ms{0.767}{0.002} \\
          & Const ($\gamma=5$)      & \bms{25.53}{0.14} / \ms{0.826}{0.001} & \ms{22.34}{0.72} / \ms{0.739}{0.002} &  \bms{26.82}{0.29} / \ms{0.838}{0.001} & \ms{14.07}{0.36} / \ms{0.728}{0.002} \\
          & Linear ($\gamma_m=5$)   & \ms{23.60}{0.13} / \ms{0.829}{0.001} & \ms{23.52}{0.14} / \bms{0.827}{0.001} &  \ms{24.14}{0.09} / \ms{0.853}{0.000} & \ms{24.09}{0.10} / \bms{0.847}{0.001} \\
          & Linear ($\gamma_m=50$)  & \ms{24.60}{0.13} / \bms{0.852}{0.001} & \ms{24.84}{0.18} / \ms{0.803}{0.001} & \ms{25.51}{0.34} / \bms{0.871}{0.001} & \ms{25.23}{0.34} / \ms{0.821}{0.000} \\
          & Linear ($\gamma_m=150$) & \ms{25.03}{0.17} / \ms{0.849}{0.001} & \bms{25.25}{0.15} / \ms{0.781}{0.001} &  \ms{26.29}{0.24} / \ms{0.867}{0.001} & \bms{25.69}{0.18} / \ms{0.781}{0.001} \\
        \midrule
        \multirow{5}{*}{100} 
          & Const ($\gamma=0.5$)    &  \ms{23.29}{0.59} / \ms{0.826}{0.001} & \ms{22.86}{0.38} / \ms{0.791}{0.002} & \ms{24.12}{0.63} / \ms{0.837}{0.001} & \ms{23.68}{0.46} / \ms{0.785}{0.002} \\
          & Const ($\gamma=5$)      & \ms{24.62}{0.44} / \ms{0.805}{0.002} &  \ms{24.05}{0.41} / \ms{0.728}{0.001} & \bms{25.58}{0.37} / \ms{0.805}{0.002}& \ms{24.43}{0.50} / \ms{0.694}{0.003} \\
          & Linear ($\gamma_m=5$)   &  \bms{23.95}{0.19} / \ms{0.839}{0.002} & \ms{23.85}{0.22} / \bms{0.833}{0.002} & \ms{24.31}{0.14} / \bms{0.854}{0.001} & \ms{24.45}{0.14} / \bms{0.851}{0.001} \\
          & Linear ($\gamma_m=50$)  &  \ms{24.49}{0.28} / \bms{0.846}{0.002} & \ms{24.37}{0.27} / \ms{0.799}{0.002} & \ms{25.07}{0.24} / \bms{0.854}{0.001} & \ms{24.54}{0.41} / \ms{0.803}{0.002} \\
          & Linear ($\gamma_m=150$) & \ms{24.56}{0.33} / \ms{0.836}{0.002} & \bms{24.44}{0.29} / \ms{0.775}{0.001} & \ms{25.34}{0.31} / \ms{0.841}{0.001} & \bms{24.56}{0.51} / \ms{0.769}{0.002} \\
          \bottomrule
    \end{tabular}
\end{table*}

In experimental settings, where forward model mismatch and measurement noise are unavoidable, a static \(\gamma\) creates a difficult sim2real trade-off. Table~\ref{tab:scheduling_results} shows that settings that work well in simulation can fail catastrophically on real data (see Const \(\gamma=5\), NFE\(=1000\)). Because simulated data exactly matches the assumed forward operator, a high constant measurement weight yields high fidelity (26.82 dB at 24 views). On real data, however, unmodeled physical effects and the approximate posterior used by diffusion-based solvers cause the same setting to accumulate errors over 1000 iterations, dropping performance to 14.07 dB due to severe global intensity drift. A scheduling strategy helps bridge this gap. Using a high \(\gamma\) in the early, high-noise stages locks the trajectory onto the correct global geometry while mismatch errors remain masked by noise. Annealing \(\gamma\) in later stages, specifically \(\gamma_t = \gamma_{\max}\frac{t}{T}\) for reverse timestep \(t \leq T\), then relaxes the measurement constraint, preventing the model from fitting experimental artifacts and allowing the prior to suppress residual noise and recover fine detail. For real data (\(\mathbf{y}_\text{exp}\)), linear decay schedules consistently give the best results across all NFEs and view counts. For example, at NFE\(=1000\) and 24 views, the linear decay schedule (\(\gamma_{\max}=150\)) reaches 25.69 dB, fully recovering from the failure of the constant schedule.

Moving further towards the requirements of experimental CT, computational feasibility becomes a major bottleneck. A key strength of DDS-like sampling schemes is their potential to maintain high reconstruction quality even with drastically reduced sampling steps. When the sampling budget is reduced to NFE=100, the scheduled approach not only remains stable but achieves the highest SSIM across all real-world configurations (0.851 at 24 views). Table \ref{tab:scheduling_results} also reveals a trade-off regarding how steep this linear decay should be. A high initial weight ($\gamma_m=150$) favors pixel-intensity accuracy (PSNR), likely because it enforces data consistency strongly in the early generative steps. Conversely, a lower weight ($\gamma_m=5$) yields the highest structural similarity (SSIM). A lower maximum weight means the generative prior has more relative influence throughout the entire process, allowing the model to enforce cleaner, structurally coherent geometric edges, even if the absolute pixel intensities drift. We must note a critical ``sweet spot'' regarding these metrics. As highlighted by the perception-distortion tradeoff \cite{blau2018perception}, visual inspection suggests that while PSNR can mask the retention of streak artifacts, SSIM might hide hallucinations by favoring plausible, piecewise-constant structures \cite{bhadra2021hallucinations, kc2026sfrc}. Consequently, despite the slightly lower numerical peaks, the intermediate schedule ($\gamma_{\max}=50$) likely offers the most well-balanced reconstructions in our preliminary experiments.

\section{Conclusion}
We have shown that although diffusion-based solvers achieve excellent reconstruction quality on idealized synthetic data, their performance degrades when moving to experimental settings. This transition introduces multiple layers of complexity, from geometric variation in the object to physical inaccuracies in the measurement process. In our controlled study, these effects fall into two main categories: distributional shift in the prior and physical mismatch in the forward model. On the prior side, the best results were obtained for the high-contrast standard SL phantom using a narrowly tailored prior. Regarding the prior, the best results were obtained on the high-contrast standard SL phantom with a narrowly tailored prior. As difficulty increased with the low-contrast design phantom, reconstruction quality generally declined. Yet a broader and more diverse prior surpassed the specialized model on this phantom, suggesting that exposure to varied structures can help even in a specific target domain. An additional shift appeared in full-view reconstructions of experimental data, which no longer exhibited piecewise-constant regions. The largest drop, however, occurred with real measurements, where forward-model mismatch introduced a trade-off: constant likelihood weighting either amplified artifacts or induced hallucinations. To address this, while also improving computational feasibility, we found that an annealed linear likelihood schedule outperformed constant weighting by emphasizing data consistency in early high-noise stages and prior guidance later. Preliminary results further suggest that this scheduled inference, also with fewer sampling steps, can match or exceed exhaustive sampling, offering a practical route toward robust experimental use.

Our study uses a canonical 2D phantom as a controlled, idealized test case. We view this as a best-case scenario, since the simple geometry allows a closely matched prior. Within this setting, we propose initial strategies to reduce the sim2real gap. In real applications, however, complex texture-rich volumes and severe noise are likely to widen this gap. Extending to 3D adds further challenges, including dependence on scan geometry, increased forward-model mismatch, and heuristic extensions of 2D diffusion models. Fully closing this gap will require further work, particularly on improved likelihood-weight scheduling and evaluation metrics that better reflect reconstruction fidelity.

\section*{Code and data availability}
Code for all experiments is available at \url{https://github.com/ezgidetu/DDS-experimentalCT}. The experimental dataset is openly available on Zenodo at \url{https://zenodo.org}~\cite{wang2026experimentalsldata}.


\bibliographystyle{IEEEtran}
\bibliography{refs}

\end{document}


\maketitle
\begin{figure*}[!htbp]
    \centering
    \includegraphics[width=.4\columnwidth,trim={4cm 13cm 4cm 13cm},clip]{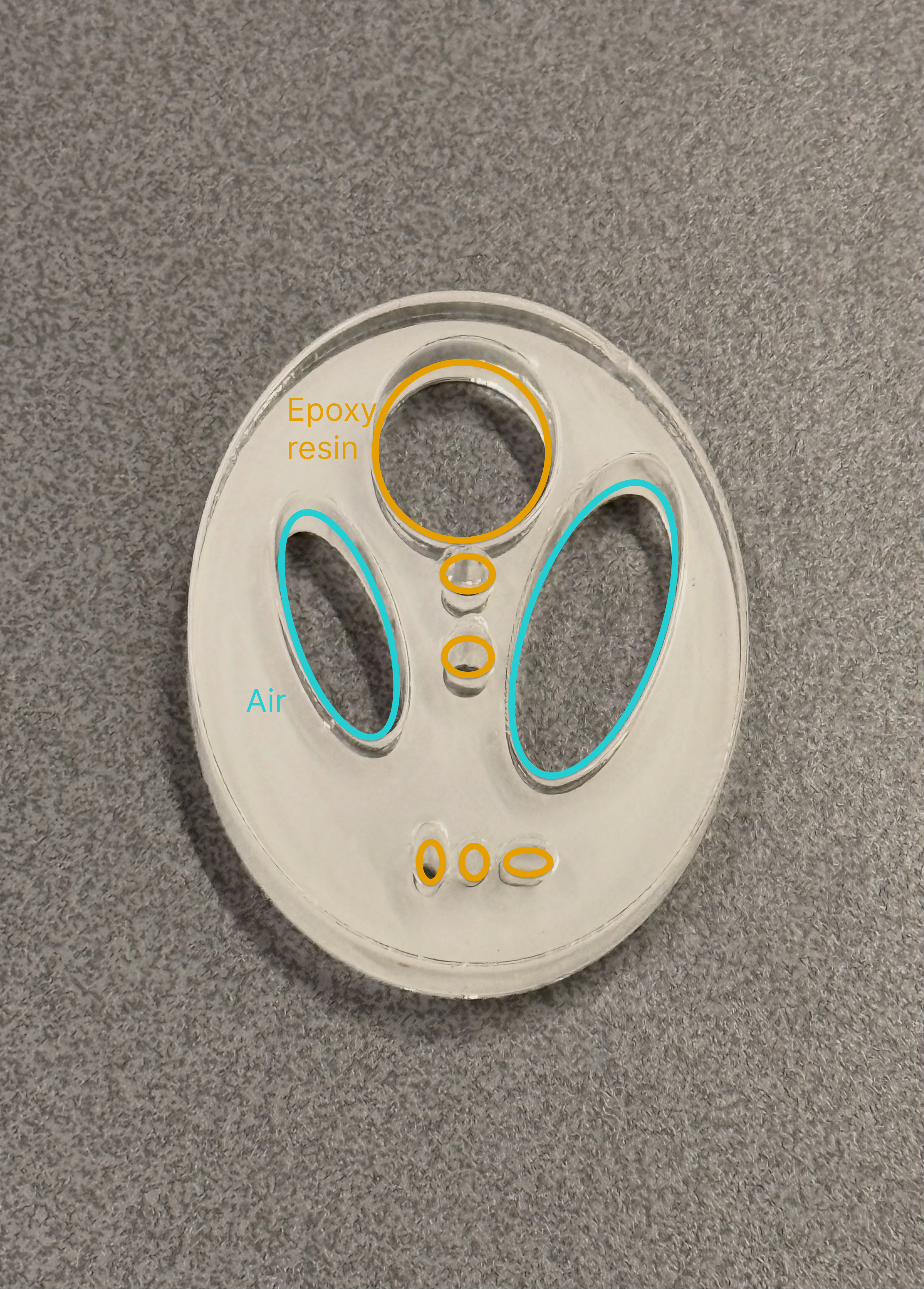}
    \caption{A photograph of the physical laser-cut phantom.}
    \label{fig:photo_physical_phantom}
\end{figure*}

\begin{figure*}[!htbp]
    \centering
    \includegraphics[width=\textwidth]{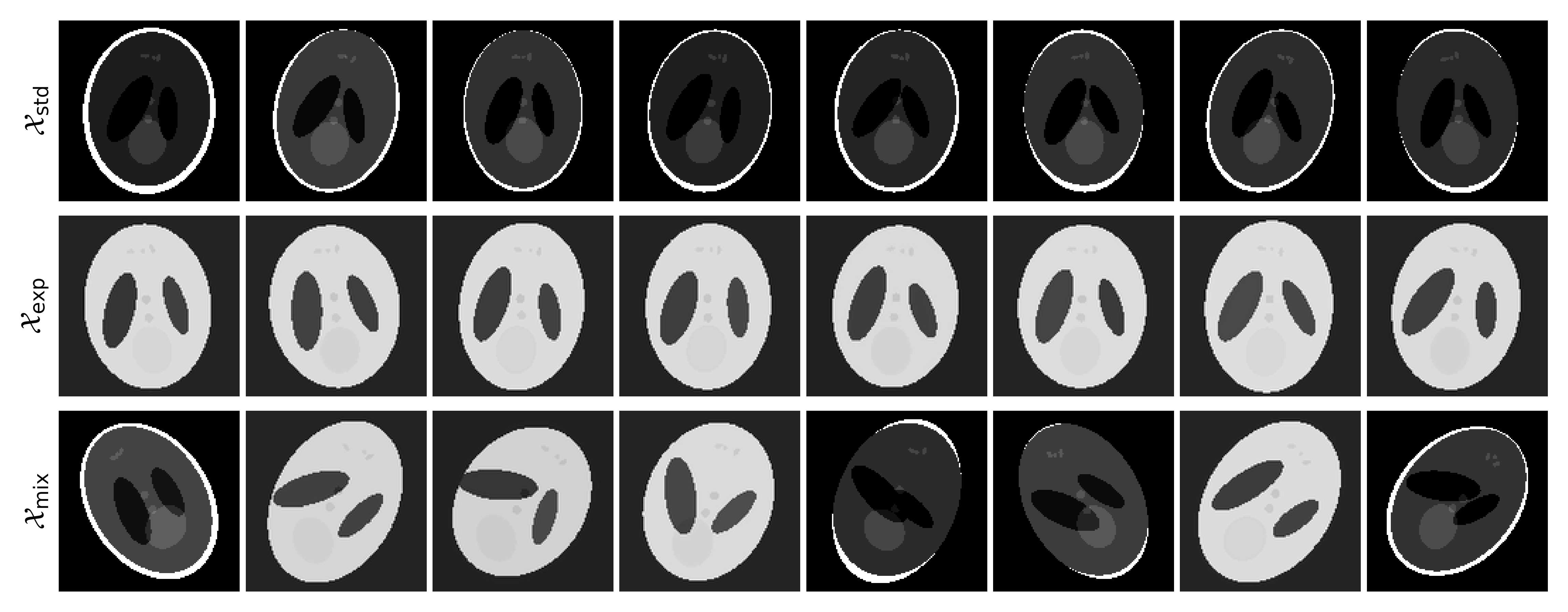}
    \caption{Randomly selected example thumbnails from three generated datasets at $128 \times 128$ resolution.}
    \label{fig:examples_from_datasets}
\end{figure*}

\begin{figure*}[!htbp]
    \centering
    \begin{subfigure}{0.95\textwidth}
        \centering
        \includegraphics[
            width=\linewidth,
            trim={0 0 0 1.2cm},
            clip
        ]{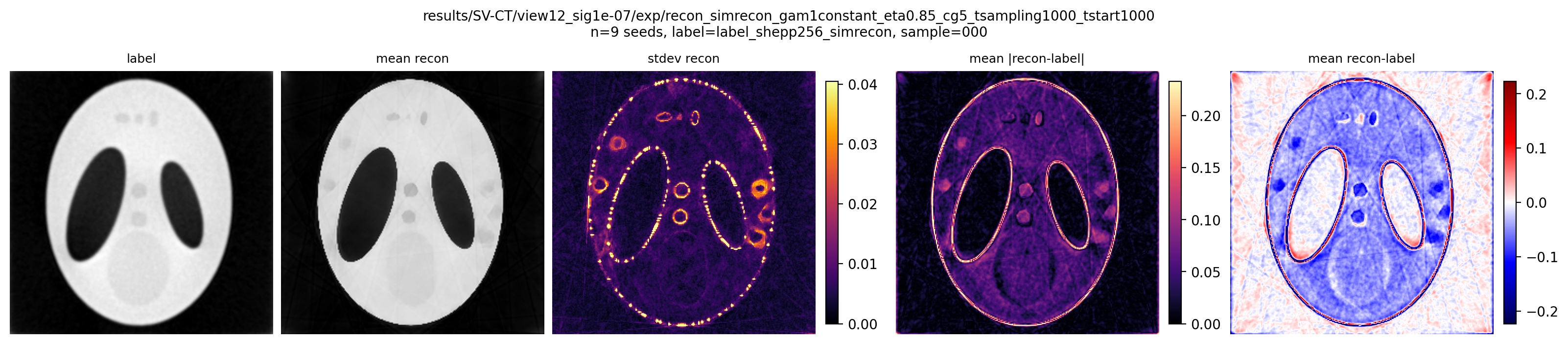}
        \caption{$\mathbf{y}_\text{sim(recon)}$, 12 projections, resolution $256 \times 256$.}
        \label{fig:uncertainty_simrecon_d12}
    \end{subfigure}
    \vspace{0.5em}
    \begin{subfigure}{0.95\textwidth}
        \centering
        \includegraphics[
            width=\linewidth,
            trim={0 0 0 1.2cm},
            clip
        ]{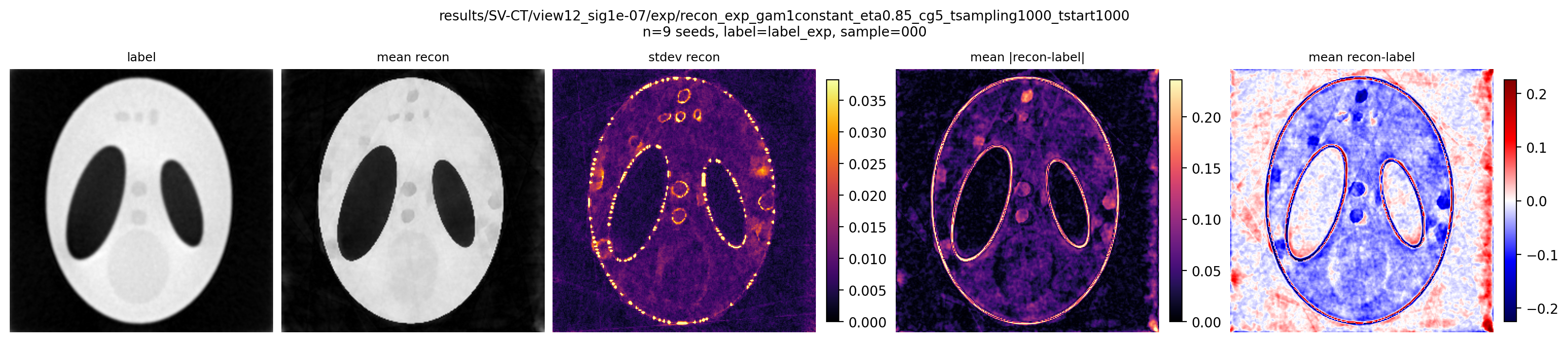}
        \caption{$\mathbf{y}_\text{exp}$, 12 projections, resolution $256 \times 256$.}
        \label{fig:uncertainty_exp_d12}
    \end{subfigure}
    \vspace{0.5em}
    \begin{subfigure}{0.95\textwidth}
        \centering
        \includegraphics[
            width=\linewidth,
            trim={0 0 0 1.2cm},
            clip
        ]{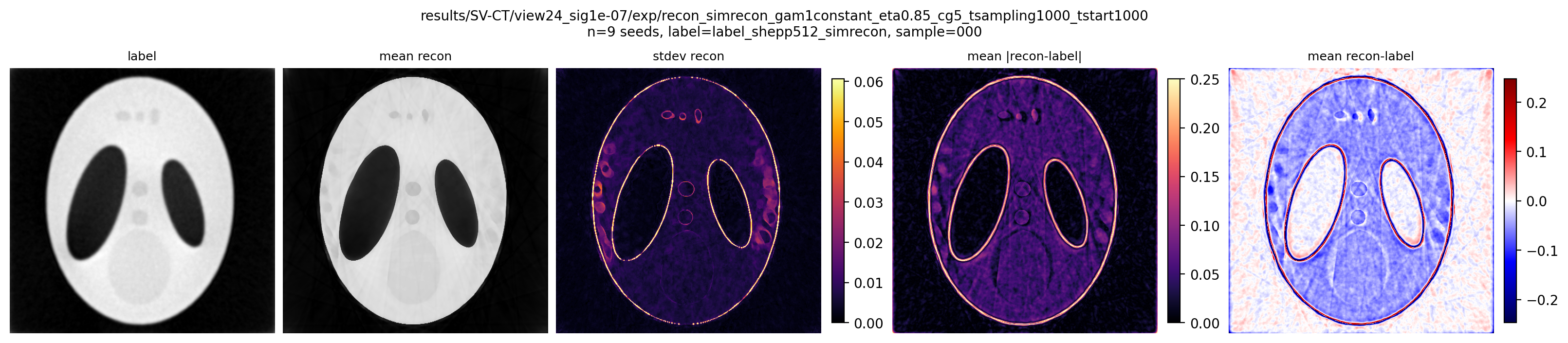}
        \caption{$\mathbf{y}_\text{sim(recon)}$, 24 projections, resolution $512 \times 512$.}
        \label{fig:uncertainty_simrecon_d24_n512}
    \end{subfigure}
    \vspace{0.5em}
    \begin{subfigure}{0.95\textwidth}
        \centering
        \includegraphics[
            width=\linewidth,
            trim={0 0 0 1.2cm},
            clip
        ]{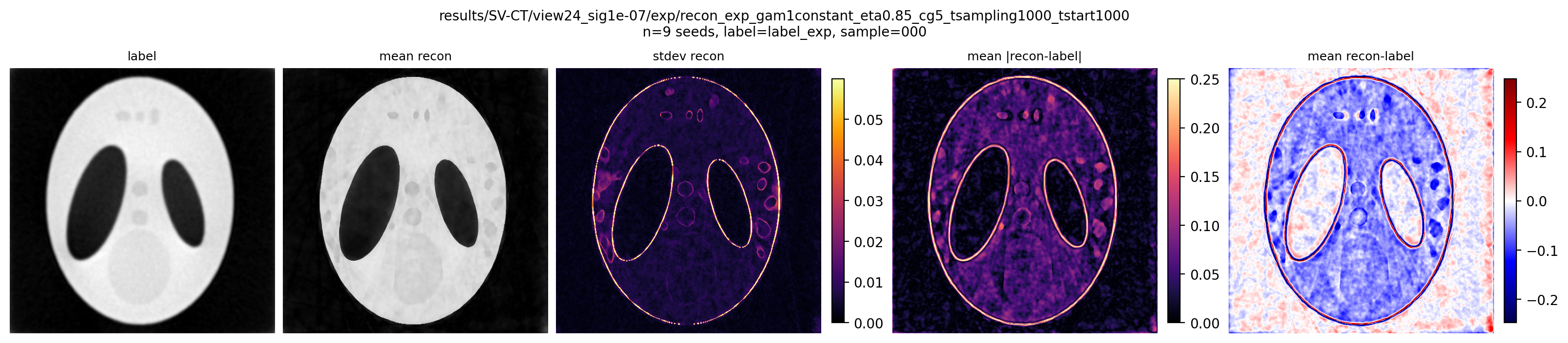}
        \caption{$\mathbf{y}_\text{exp}$, 24 projections, resolution $512 \times 512$.}
        \label{fig:uncertainty_exp_d24_n512}
    \end{subfigure}

    \caption{Reconstructions averaged over nine samples and corresponding uncertainty summaries: standard deviation, mean absolute difference, and mean difference (columns 3–5, respectively), for $\mathbf{y}_\text{sim(recon)}$ and $\mathbf{y}_\text{exp}$ at resolutions of $256 \times 256$ and $512 \times 512$, using $\mathbf{f}_\text{exp}$ as diffusion prior and under the same experimental setup with Fig. 2.}
    \label{fig:examples_from_datasets}
\end{figure*}

\begin{figure*}[!htbp]
    \centering
    \includegraphics[width=.85\textwidth]{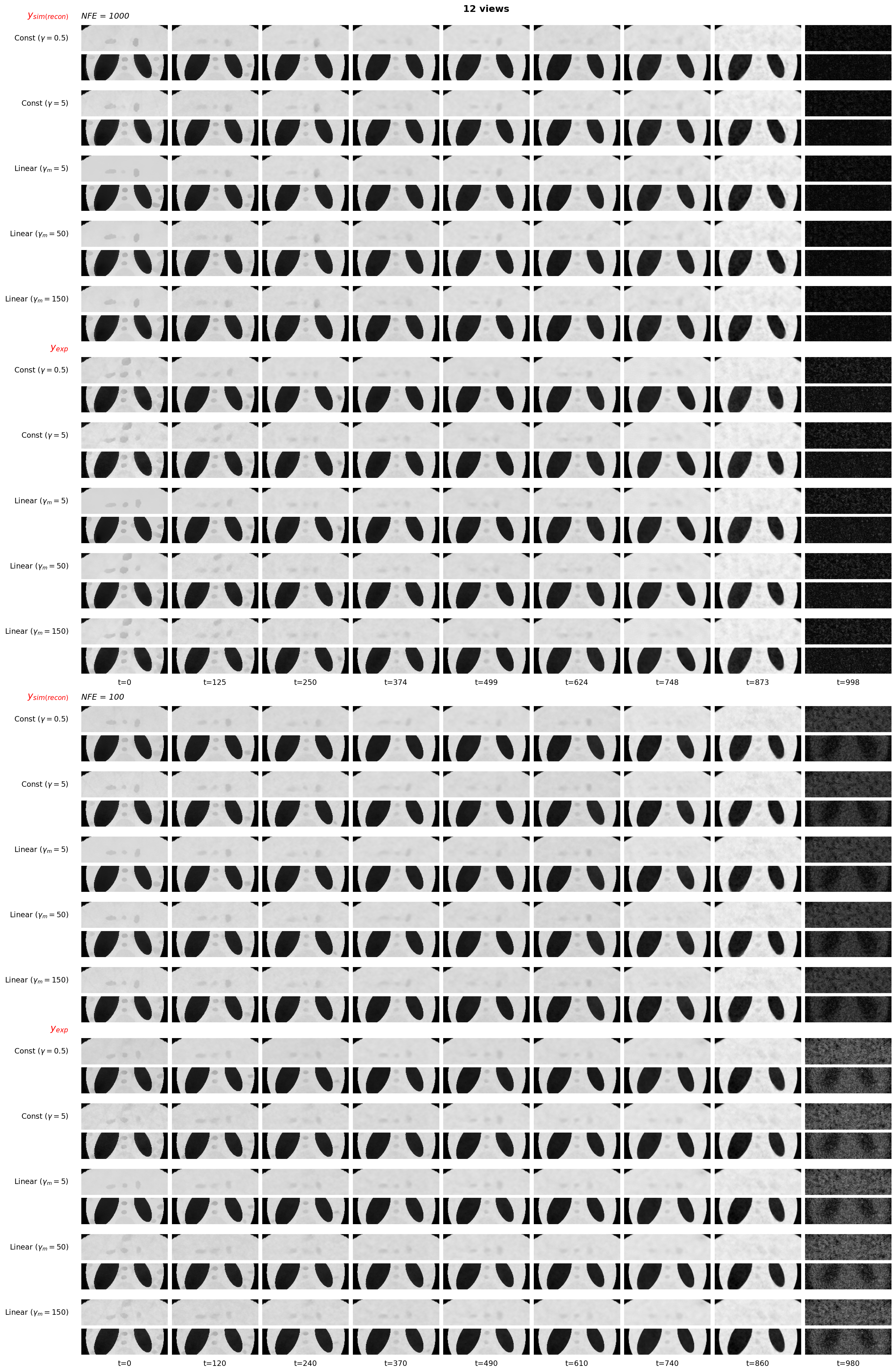}
    \caption{Reconstruction trajectories for the cases shown in Table~II.}
    \label{fig:examples_from_datasets}
\end{figure*}

\begin{figure*}[!htbp]
    \centering
    \includegraphics[width=.85\textwidth]{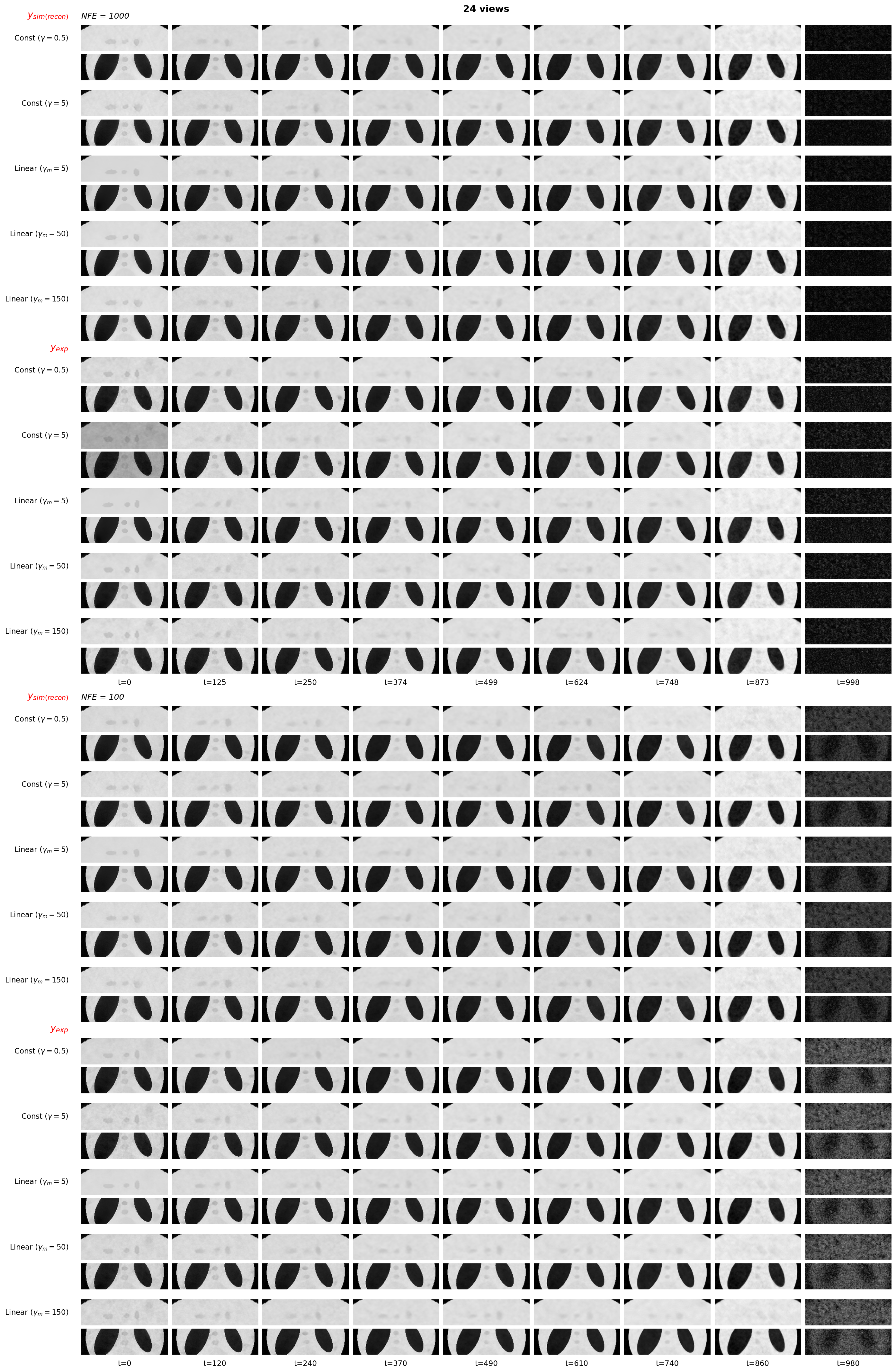}
    \caption{Reconstruction trajectories for the cases shown in Table~II (continued).}
    \label{fig:examples_from_datasets}
\end{figure*}

\begin{table*}[t!]
    \centering
    \newcommand{\bms}[2]{$\mathbf{#1}_{\scriptscriptstyle #2}$}
    \newcommand{\ms}[2]{$#1_{\scriptscriptstyle #2}$}
    \caption{Continued quantitative comparison (PSNR / SSIM) of constant vs linearly decaying likelihood-weight schedules at a $256 \times 256$ resolution, focusing on highly reduced sampling steps (NFE=10), for both $\mathbf{y}_{\text{sim(recon)}}$ and $\mathbf{y}_{\text{exp}}$.}
    \label{tab:scheduling_results}
    \setlength{\tabcolsep}{3pt} 
    \begin{tabular}{c l c c c c}
        \toprule
        & & \multicolumn{2}{c}{\textbf{12 views}} & \multicolumn{2}{c}{\textbf{24 views}} \\
        \cmidrule(lr){3-4} \cmidrule(lr){5-6}
        \textbf{NFE} & \textbf{Schedule} & $\mathbf{y}_\text{sim(recon)}$ & $\mathbf{y}_\text{exp}$ & $\mathbf{y}_\text{sim(recon)}$ & $\mathbf{y}_\text{exp}$ \\
        \midrule
        \multirow{5}{*}{10} 
          & Const ($\gamma=0.5$)    & \ms{20.30}{0.93} / \ms{0.757}{0.024} & \ms{20.00}{0.81} / \ms{0.736}{0.020} & \ms{21.30}{0.70} / \ms{0.787}{0.015} & \ms{21.28}{0.88} / \ms{0.762}{0.014} \\
          & Const ($\gamma=5$)      & \ms{22.23}{0.52} / \ms{0.759}{0.013} & \ms{21.67}{0.43} / \ms{0.715}{0.011} & \ms{22.96}{0.65} / \ms{0.777}{0.005} & \ms{22.22}{0.45} / \ms{0.707}{0.005} \\
          & Linear ($\gamma_m=5$)   & \ms{22.66}{0.49} / \ms{0.785}{0.007} & \ms{22.69}{0.30} / \ms{0.791}{0.007} & \ms{23.44}{0.24} / \ms{0.816}{0.003} & \bms{23.08}{0.42} / \ms{0.813}{0.006} \\
          & Linear ($\gamma_m=50$)  & \bms{22.86}{0.54} / \bms{0.810}{0.005} & \bms{22.97}{0.26} / \bms{0.797}{0.008} & \bms{23.64}{0.47} / \bms{0.832}{0.003} & \ms{23.36}{0.41} / \bms{0.814}{0.006} \\
          & Linear ($\gamma_m=150$) & \ms{22.84}{0.56} / \ms{0.809}{0.006} &  \ms{22.85}{0.26} / \ms{0.787}{0.009} & \ms{23.57}{0.43} / \ms{0.830}{0.003} & \ms{23.29}{0.35} / \ms{0.803}{0.006} \\
        \bottomrule
    \end{tabular}
\end{table*}